\documentclass[twocolumn,10pt,cleanfoot,cleanhead,notitlepage]{asme2e}

\usepackage{graphicx}
\usepackage{amsmath}
\confshortname{IDETC/CIE2021}
\conffullname{the ASME 2021 \\ International Design Engineering Technical Conferences and\\
              Computers and Information in Engineering Conference}

\confdate{August 17-20}
\confyear{2021}
\confcity{Virtual}
\confcountry{Online}

\papernum{DETC2021-68093}

\title{Design And Analysis Of Three-Output Open Differential with 3-DOF}

\author{Rama Vadapalli, Nagamanikandan Govindan, K Madhava Krishna
    \affiliation{International Institute of Information Technology Hyderabad, Hyderabad, India}
}

\begin{document}

\maketitle    

\begin{abstract}
{\it This paper presents a novel passive three-output differential with three degrees of freedom (3DOF), that translates motion and torque from a single input to three outputs. The proposed Three-Output Open Differential is designed such that its functioning is analogous to the functioning of a traditional two-output open differential. That is, the differential translates equal motion and torque to all its three outputs when the outputs are unconstrained or are subjected to equivalent load conditions. The introduced design is the first differential with three outputs to realise this outcome. The differential action between the three outputs is realised passively by a symmetric arrangement of three two-output open differentials and three two-input open differentials. The resulting differential mechanism achieves the novel result of equivalent input to output angular velocity and torque relations for all its three outputs. Furthermore, Three-Output Open Differential achieves the novel result for differentials with more than two outputs where each of its outputs shares equivalent angular velocity and torque relations with all the other outputs. The kinematics and dynamics of the Three-Output Open Differential are derived using the bond graph method. In addition, the merits of the differential mechanism along with its current and potential applications are presented.}
\end{abstract}

\section*{INTRODUCTION}

A differential is a mechanism that translates motion and torque from an input to its outputs in accordance with the load conditions the outputs experience \cite{alcacer2015design}. The traditional two-output open differential $(2-OD)$ has a single input which passively operates two outputs. It has two degrees-of-freedom (2DOF) and when in operation, translates equal motion and torque from its single input to its two outputs if they are subjected equal loads \cite{chu2013conceptual}. However, when the load conditions of the outputs differ, the motion translated to each output is different. The $2-OD$ plays a vital role in automobiles. During a turn, the $2-OD$ enables the automobile to rotate the wheel that travels a longer distance faster than the other wheel and potentially eliminates the slip and drag of the wheels \cite{uicker2003theory}. By doing so, the $2-OD$ considerably reduces the stresses induced. Three-output differentials play a similar role in delivering motion to three outputs.

\begin{figure} 
\centerline{\includegraphics[width=3.34in]{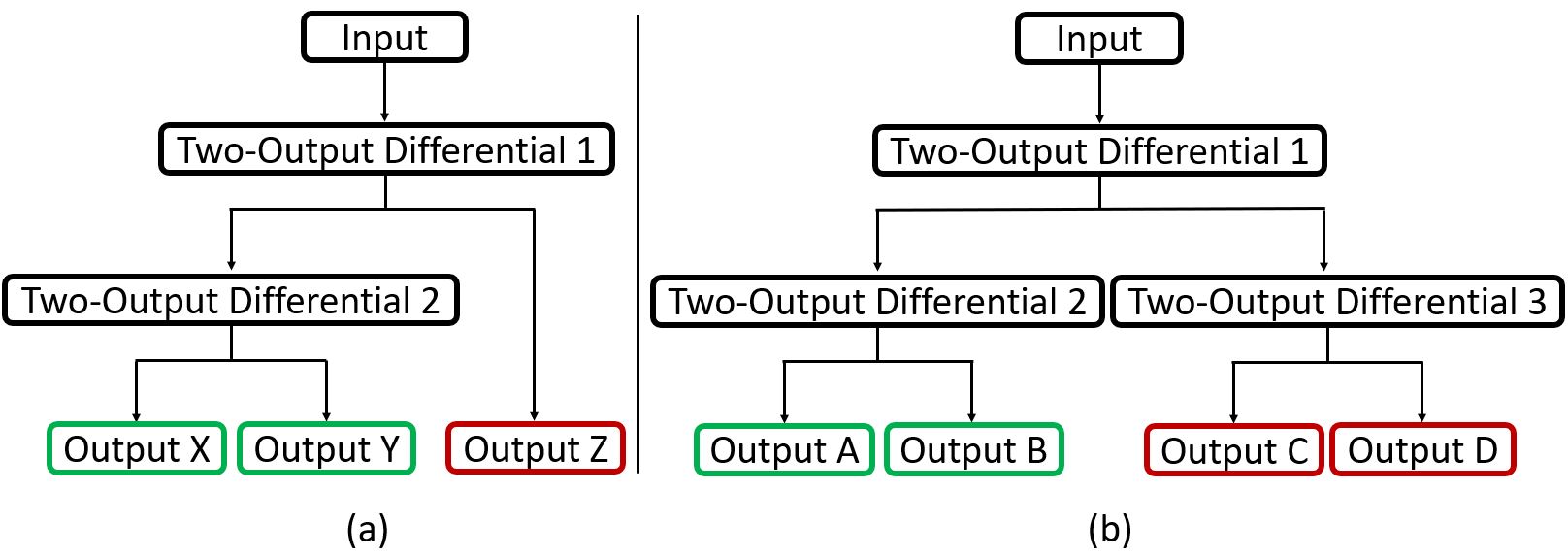}}
\caption{CURRENT DIFFERENTIALS WITH (a) THREE OUTPUTS (b) FOUR OUTPUTS.}
\label{figure1}\vspace{-0.2in}
\end{figure}

Three-output differentials have existed for over a decade and have predominantly been used in pipe climbing robots with three tracks or wheels \cite{kim2013pipe, yang2014novel, kim20152, kim2016novel}. One such differential, the multi-axle differential gear was implemented in the in-pipe robot MRINSPECT VI (Multifunctional Robotic crawler for In-pipe inspection VI) \cite{yang2014novel}. The multi-axle differential gear has three sets of planetary gears arranged in a series where its three outputs are the ring gears of each planetary gear. The series arrangement of the planetary gears permit the ring gears to rotate at different speeds, however; it encounters a limitation \cite{kim2016novel}. The input does not translate equal motion to the outputs when the outputs are subjected to equal loads. This limitation is transpired because all three outputs do not have equivalent dynamics with the input. The schematic of the mechanism is as shown in Fig.~\ref{figure1}$(a)$ where one of the outputs $($Output $Z)$ has a different link to the input when compared to the other two outputs $($Output $X$ and Output $Y)$. This causes the three outputs of the differential to behave differently under equal loads \cite{kim2016novel}.

Another solution for three-output differential was developed by S. Kota and S. Bidare $(1997)$ using standard bevel gear setups and epicyclic gear trains $($planetary gear$)$ \cite{kota1997systematic}. A similar approach is proposed by Diego Ospina and Alejandro Ramirez-Serrano $(2020)$ for the use of differentials with more than 2DOF in underactuated robotic arms \cite{ospina2020sensorless}. The differentials proposed by S. Kota and S. Bidare $(1997)$ and by Diego Ospina and Alejandro Ramirez-Serrano $(2020)$ have schematics similar to those shown in Fig.~\ref{figure1}$(a)$ and carry similar limitations \cite{kota1997systematic, ospina2020sensorless}. 

There exist four-output open differentials such as the 2-2D differential gear mechanism that translates equal motion to all its outputs when the outputs are under equivalent load conditions \cite{kim20152, kim2016novel}. The 2-2D differential has four sets of planetary gears arranged in a series where its four outputs are the ring gears of each planetary gear. Its schematic Fig.~\ref{figure1}$(b)$ is such as connecting the inputs of two $2-OD$ to the outputs of another $2-OD$ \cite{kim2016novel}. The 2-2D differential gear mechanism has a limitation, that is the effect an output has on the other three outputs is not equivalent even when the other outputs are under the equal loads. This transpires because the dynamics an output shares with the other three outputs is not equivalent. This limitation becomes more apparent when one of outputs’ (Output A) motion is ceased and the other three outputs (Output B, Output C and Output D) are unconstrained Fig.~\ref{figure1}$(b)$. In such cases the Output B will rotate faster than the Output C and Output D.

In addition to using differential mechanisms, differential speed can be realised by electronically controlling the speeds of each wheel. Our previously developed robots, the Modular Pipe Climber I and II predefined the speeds for each track in accordance to the bends of the pipe \cite{vadapalli2019modular, suryavanshi2019omnidirectional, agarwal2021design}. Other pipe climbing robots such as MRINSPECT $IV$ and LS-$01$ have also used similar methods in modulating speeds \cite{roh2005differential, baharuddin2012robot}. The need to pre-define the speeds of each track is a substantial limitation which restricts the robots’ success to a highly predictable environment.
\begin{figure} 
\centerline{\includegraphics[width=2.9in]{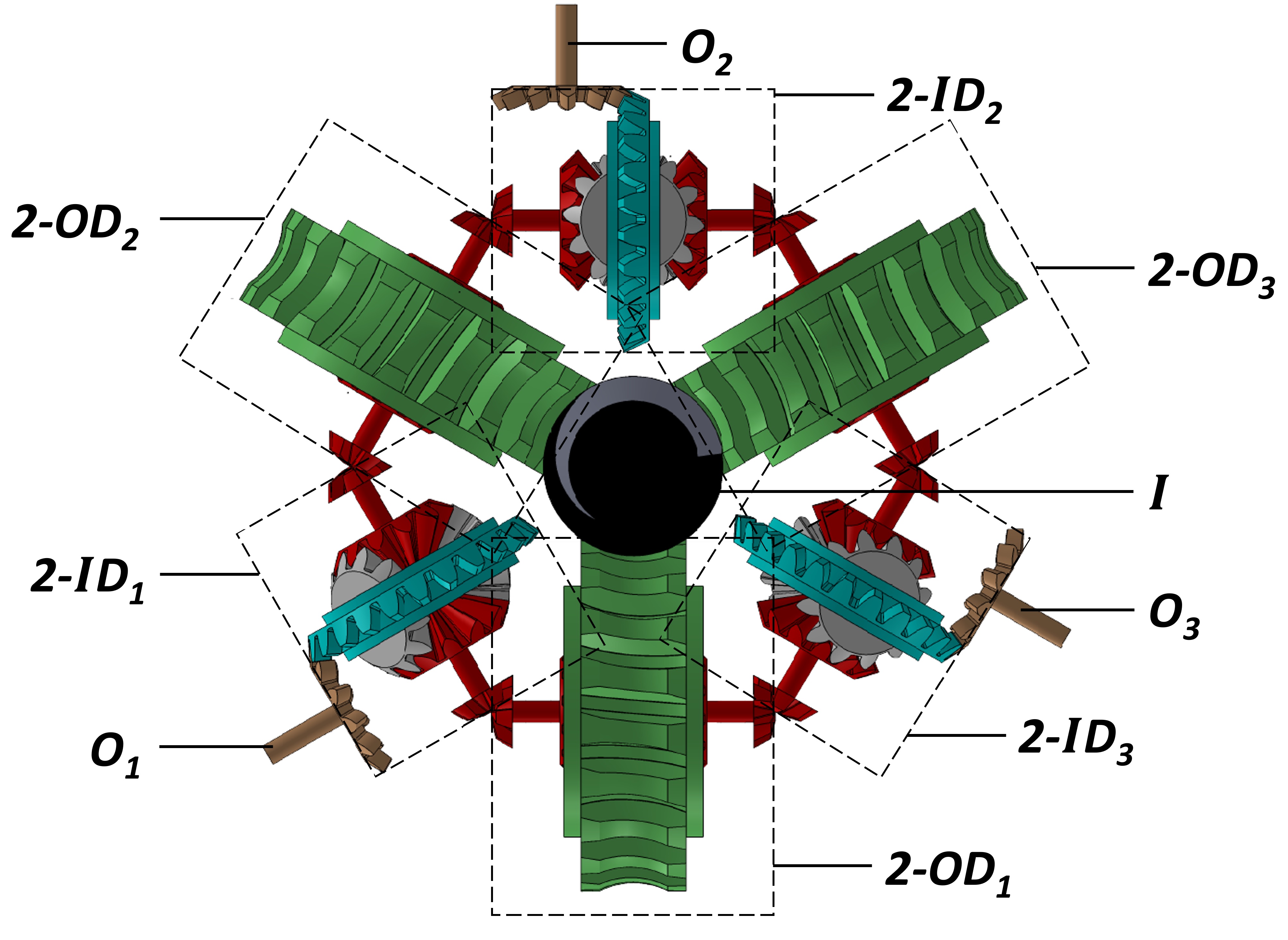}}
\caption{THREE-OUTPUT OPEN DIFFERENTIAL $(3-OOD)$.}
\label{figure2}\vspace{-0.2in}
\end{figure}

\textbf{Contribution:} The Three-Output Open Differential $(3-OOD)$ as shown in Fig.~\ref{figure2} is designed such that it realises all the functional abilities of a traditional $2-OD$, but with three outputs, i.e., $(i)$ the three outputs operate with equal speed when the outputs are unconstrained or under equal loads, $(ii)$ all three outputs have equivalent input to output kinematics and dynamics and $(iii)$ the kinematics and dynamics between any two of its outputs is equivalent.

The design of the $3-OOD$ along with its components are presented in the next section. The section (Kinematics and Dynamics), establishes the kinematics and dynamics of the $3-OOD$ by means of the bond graph method. Finally, the current and potential applications of the mechanism along with our future work are presented.

\section*{DESIGN OF THE 3-OOD}
\vspace{0.0in}
The design intent for the $3-OOD$ was to develop a functional architecture that could achieve all the results achieved by the traditional $2-OD$, but with three outputs.
\subsection*{Design Synthesis}
\vspace{0.0in}

The functioning of the $2-OD$ and the principles it operates are studied for the purpose of incorporating these principles into the design of the $3-OOD$. $2-OD$ chiefly operates on two principles. (i) The effect the input has on its outputs is variable with the load conditions the outputs experience \cite{hsu1998methodology}. (ii) The input shares equivalent dynamics with all its outputs \cite{hsu1998methodology}. However, there is a third principle that needs to be satisfied when the differential has more number than two outputs, that is, (iii) the dynamics an output shares with all the other outputs is equivalent. The currently existing open differentials with three outputs as represented in Fig.~\ref{figure1}$(a)$ only satisfy the first principle. The currently existing open differentials with four outputs as represented in Fig.~\ref{figure1}$(b)$ satisfy the first principle and second principle, but do not operate under the third principle.
\begin{figure} 
\centerline{\includegraphics[width=3.25in]{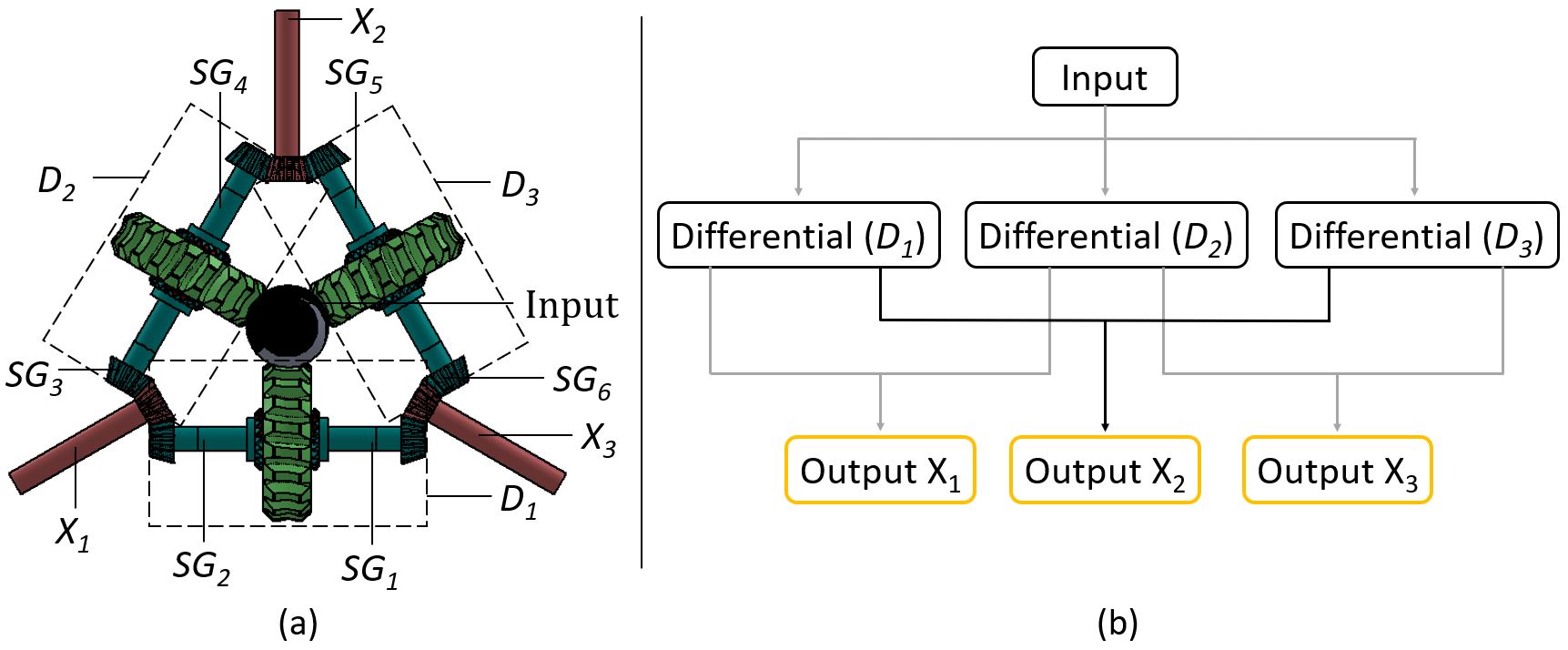}}
\caption{INITIAL DESIGN (a) TOP VIEW (b) SCHEMATIC.}
\label{figure3}\vspace{-0.2in}
\end{figure}

To integrate all the three principles determined into a design, the initial synthesis was to have a single central input transfer motion to three $2-OD$'s $(D_1$, $D_2$ and $D_3)$ as shown in Fig.~\ref{figure3}$(a)$. Each side gear $(SG_1$, $SG_2$, $SG_3$, $SG_4$, $SG_5$ and $SG_6)$ of the three $2-OD$'s is connected with the adjacent side gear of its neighbouring $2-OD$ to combine and form an output $(X_1$, $X_2$ and $X_3)$ of a three-output differential Fig.~\ref{figure3}$(a)$. Figure~\ref{figure3}$(b)$ illustrates the schematic of the initial design. On analysing the initial design, it was observed that though the initial design could achieve differential speed for its three outputs, the two adjacent side gears $(SG_2$, $SG_3)$ or $(SG_4$, $SG_5)$ or $(SG_1$, $SG_6)$ at the respective outputs $(X_1$, $X_2$ and $X_3)$ are forced to move with equal speeds. As a consequence, at any given moment, at least two of the three outputs will always have equal speeds regardless of the load conditions the outputs experiences.
\begin{figure} 
\centerline{\includegraphics[width=2.9in]{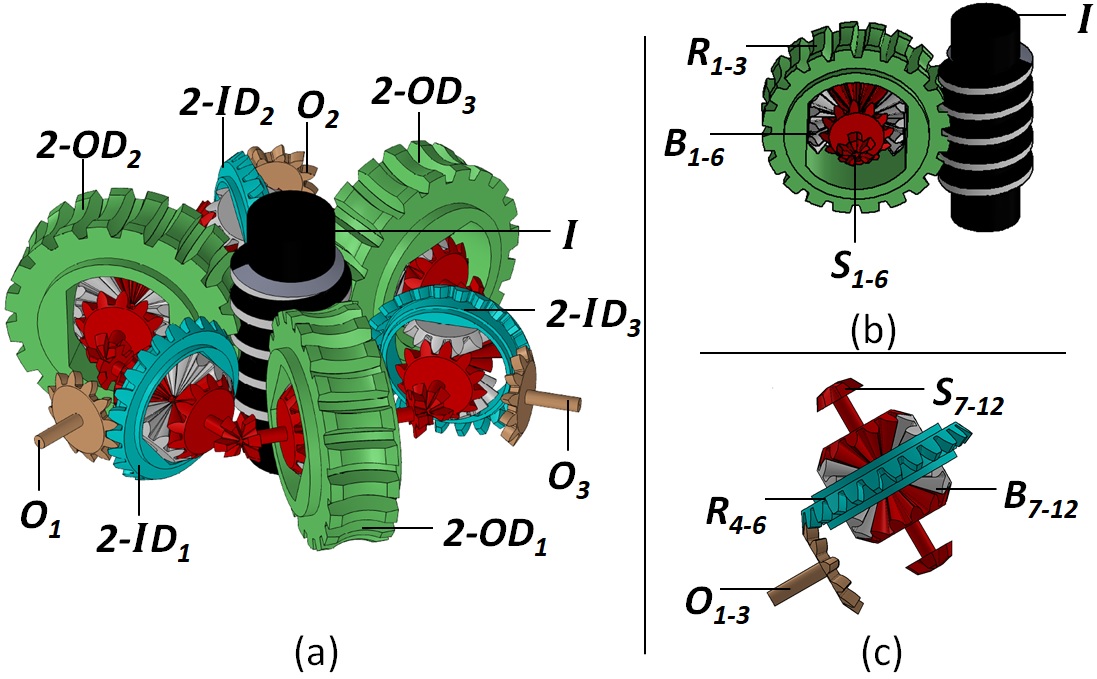}}
\caption{(a) $3-OOD$ (b) TWO-OUTPUT DIFFERENTIAL $(2-OD_{1-3}$) (C) TWO-INPUT DIFFERENTIAL $(2-ID_{1-3})$.}
\label{figure4}\vspace{-0.2in}
\end{figure}

To eliminate the limitation present in the initial design three two-input open differentials $(2-ID_1$, $2-ID_2$ and $2-ID_3)$ are used so that they facilitate differential speed between the adjacent side gears at the outputs as shown in Fig.~\ref{figure4}. Each two-input differential receives motion from the adjacent side gears of its neighbouring two-output differentials $(2-OD_1$, $2-OD_2$ and $2-OD_3)$ and combines the motion to a single output $(O_1$, $O_2$ and $O_3)$ as shown in Fig.~\ref{figure4}.
\begin{figure} 
\centerline{\includegraphics[width=2.9in]{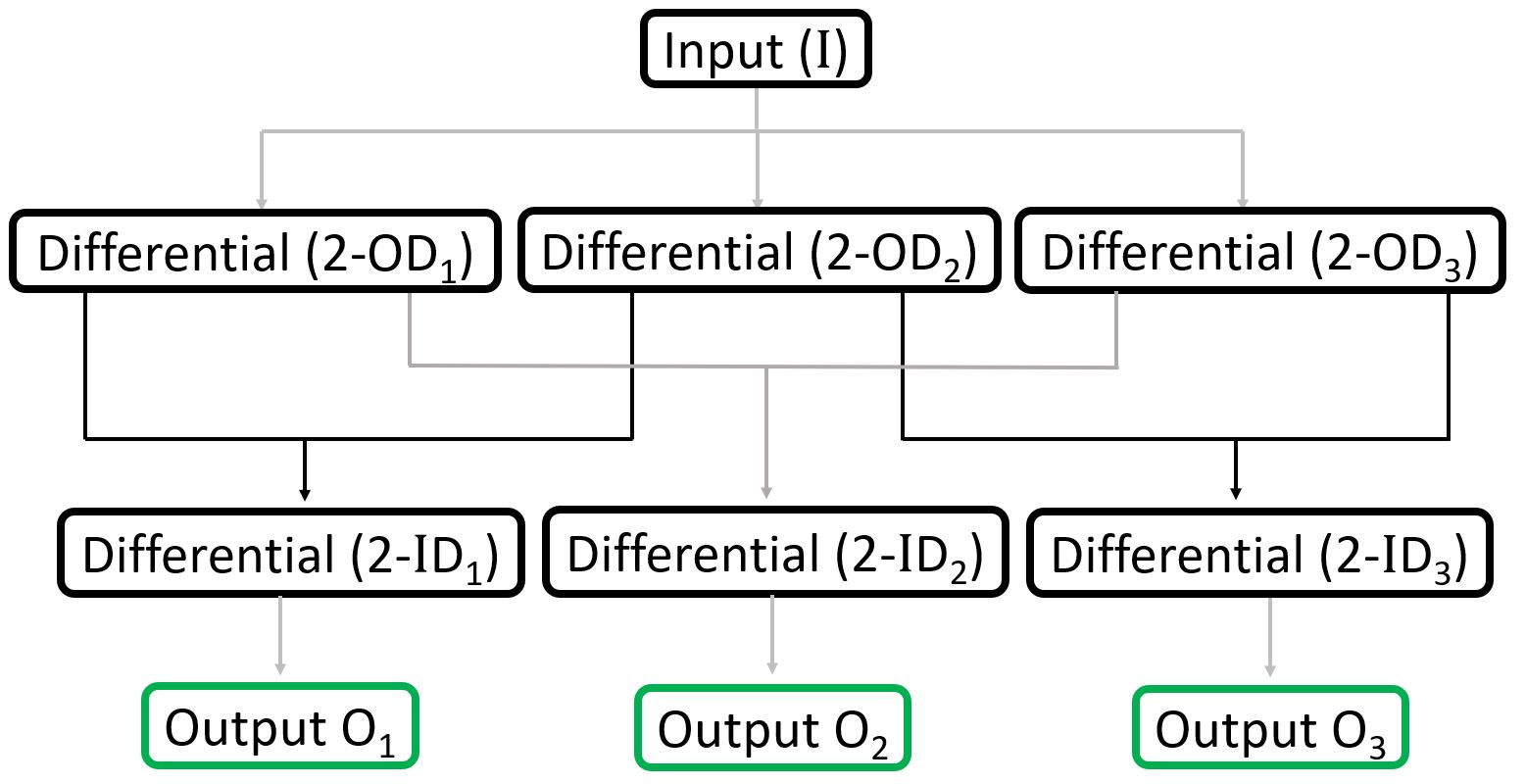}}
\caption{SCHEMATIC OF THE $3-OOD$.}
\label{figure5}\vspace{-0.2in}
\end{figure}

\textbf{\small Novelty of \textit{3-OOD}:\normalsize} The proposed design of the $3-OOD$ satisfies all the three aforementioned principles of operation of the traditional $2-OD_2$. The schematic as shown in Fig.~\ref{figure5} illustrates that each output shares the same dynamics with the other outputs. As a result, the change of loads for one of the outputs will have the equal effect on the remaining outputs.

\subsection*{Components Of The Differential}
\vspace{0.0in}

\begin{figure} 
\centerline{\includegraphics[width=3in]{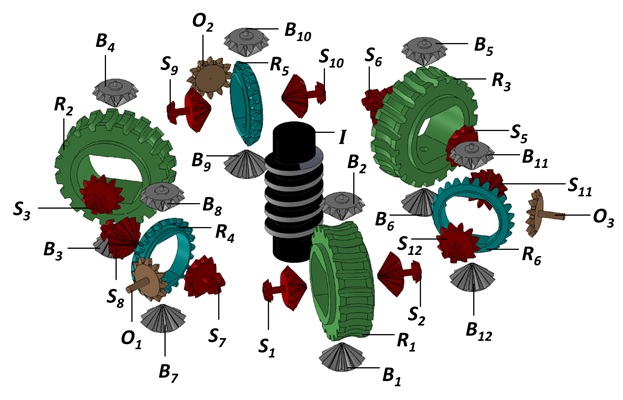}}
\caption{$3-OOD$ (EXPLODED VIEW).}
\label{figure6}\vspace{-0.2in}
\end{figure}

The design of the $3-OOD$ comprises a single input $(I)$, three two-output open differentials $(2-OD_{1-3})$, three two-input open differentials $(2-ID_{1-3})$ and three outputs $(O_{1-3})$ as shown in Fig.~\ref{figure4} and Fig.~\ref{figure6}. The input of the $3-OOD$ is located at the centre of the mechanism and the three two-output differentials $(2-OD_{1-3})$ are arranged symmetrically around the input in a circular pattern at $120^o$ apart from each other. Each of the three two-input differential $(2-ID_{1-3})$ is arranged radially in between the two-output differentials. The three two-input differentials form a circular pattern with $120^o$ between each other and bisect the two-output differentials with an angle of $60^o$ as shown in Fig.~\ref{figure2}. The outputs of the three two-input differentials $(2-ID_{1-3})$ form to be the three outputs of the $3-OOD$.

Each of the three $2-OD_{1-3}$ have six moving parts, an input $(I)$ which is shared by all three, a ring gear $(R_{1-3})$ that is connected to the input, two bevel gears $(B_{1-6})$ which are arranged inside the ring gear and two side gears $(S_{1-6})$, which are connected to its bevel gears Fig.~\ref{figure4} and Fig.~\ref{figure6}.  Similarly, each of the three $2-ID_{1-3}$ have six moving parts, an output $(O_{1-3})$, a ring gear $(R_{4-6})$ that is connected to the output, two bevel gears $(B_{7-12})$ which are arranged inside the ring gear and two side gears $(S_{7-12})$, which are connected to its bevel gears Fig.~\ref{figure4}.

The input $(I)$ provides motion to the three ring gears $(R_{1-3})$ of $2-OD_{1-3}$. $2-OD_{1-3}$ translate the motion received by their respective ring gears $(R_{1-3})$ to their respective side gears $(S_{1-6})$ depending on the load conditions the side gears experience. The side gears $(S_{1-6})$ transfer their motion to the adjoining side gears $(S_{7-12})$ of $2-ID_{1-3}$, which is translated to the three outputs $(O_{1-3})$ via their respective ring gears $(R_{4-6})$. When the two side gears of a two-input differentials ($2-ID_{1-3}$) move with equal speeds, their respective bevel gears do not engage, and the ring gear will move with the same speed as the side gears. When the two side gears of a two-input differential ($2-ID_{1-3}$) have different speeds, $2-ID_{1-3}$ translates the differential speed to a single output. Under equal loads, all the twelve side gears $(S_{1-12})$ and the six ring gear $(R_{1-6})$ move at equal speeds. Overall, the design of the $3-OOD$ comprises six open differentials symmetrically arranged radially at $60^o$ to each other. The six differentials work together to translate the motion from the ring gears of the three two-output differentials $(2-OD_{1-3})$ to the ring gears of the three two-input differentials $(2-ID_{1-3})$. The three ring gears $(R_{4-6})$ of the three two-input differentials independently power the three outputs $(O_{1-3})$. The $3-OOD$ is symmetric in design with the centre of mass of the mechanism coinciding with its geometric centre, which ensures that the mechanism is well balanced around its geometric centre in terms of weight and vibrations that can occur during its operation.

In total, the $3-OOD$ has thirty-four moving parts, a single input $(I)$, three outputs $(O_{1-3})$, six ring gears $(R_{1-6})$, twelve bevel gears $(B_{1-12})$ and twelve side gears $(S_{1-12})$ as shown in Fig.~\ref{figure6}. The gear arrangement between the central input and the ring gear of each of $2-OD_{1-3}$ is a worm gear arrangement. However; the central input and the ring gears of $2-OD_{1-3}$ can also be configured to a crown gear arrangement. Each of the twelve bevel gears have a crown gear and each of the twelve side gears have a crown gear at one end of the shaft which connect to the bevel gears of its respective differential. The side gears have another crown gear at the other end of the shaft that is designed to operate at an angle of $120^o$ and is connected to side gear of the adjacent differential as shown in Fig.~\ref{figure4} and Fig.~\ref{figure6}. The gear arrangement between the outputs and the ring gears of $2-ID_{1-3}$ is a crown gear arrangement.

\section*{KINEMATICS AND DYNAMICS}

\begin{figure} 
\centerline{\includegraphics[width=3.34in]{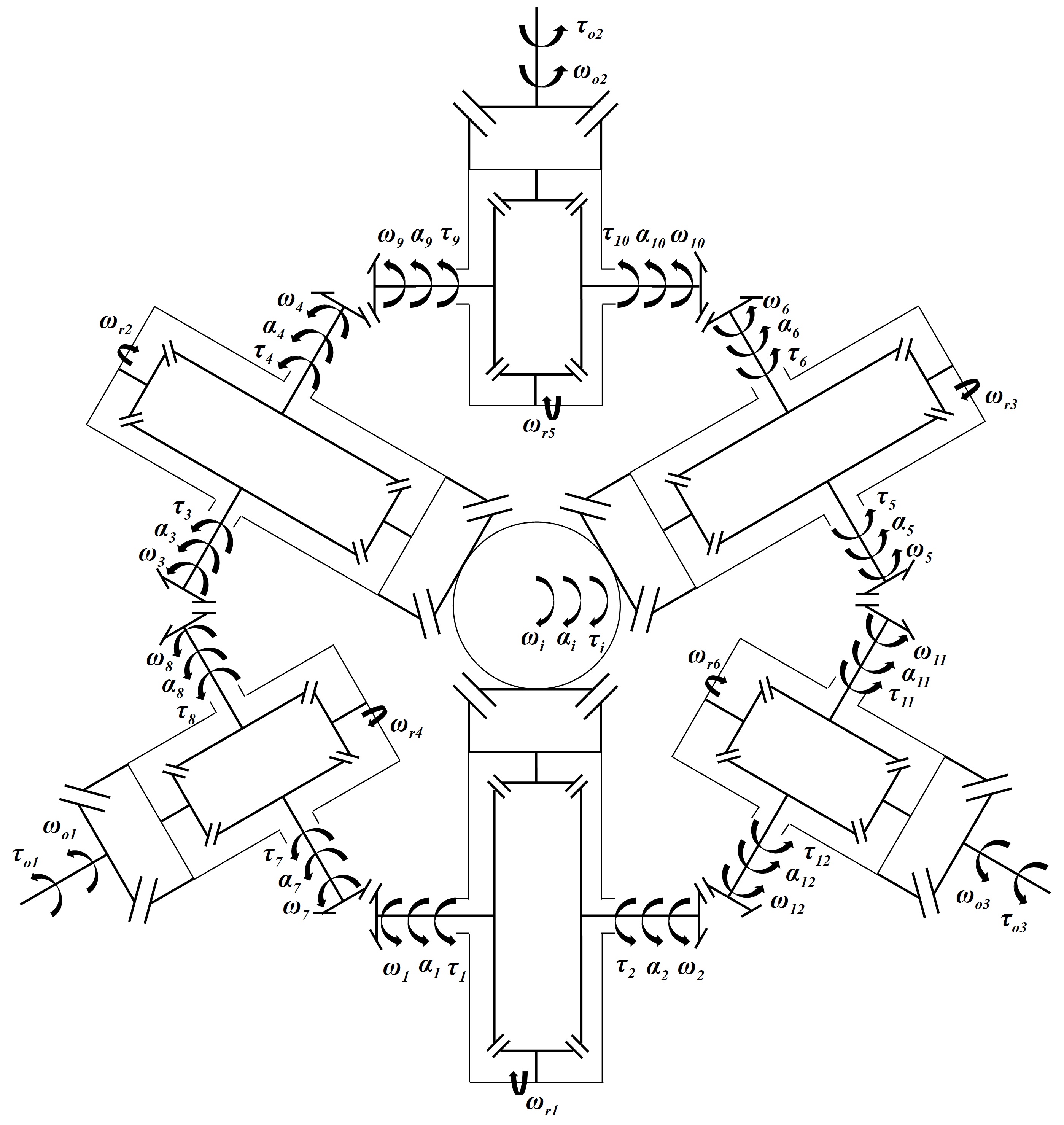}}
\caption{KINEMATIC SCHEME OF THE DIFFERENTIAL.}
\label{figure7}\vspace{-0.15in}
\end{figure}

\begin{figure} 
\centerline{\includegraphics[width=3.34in]{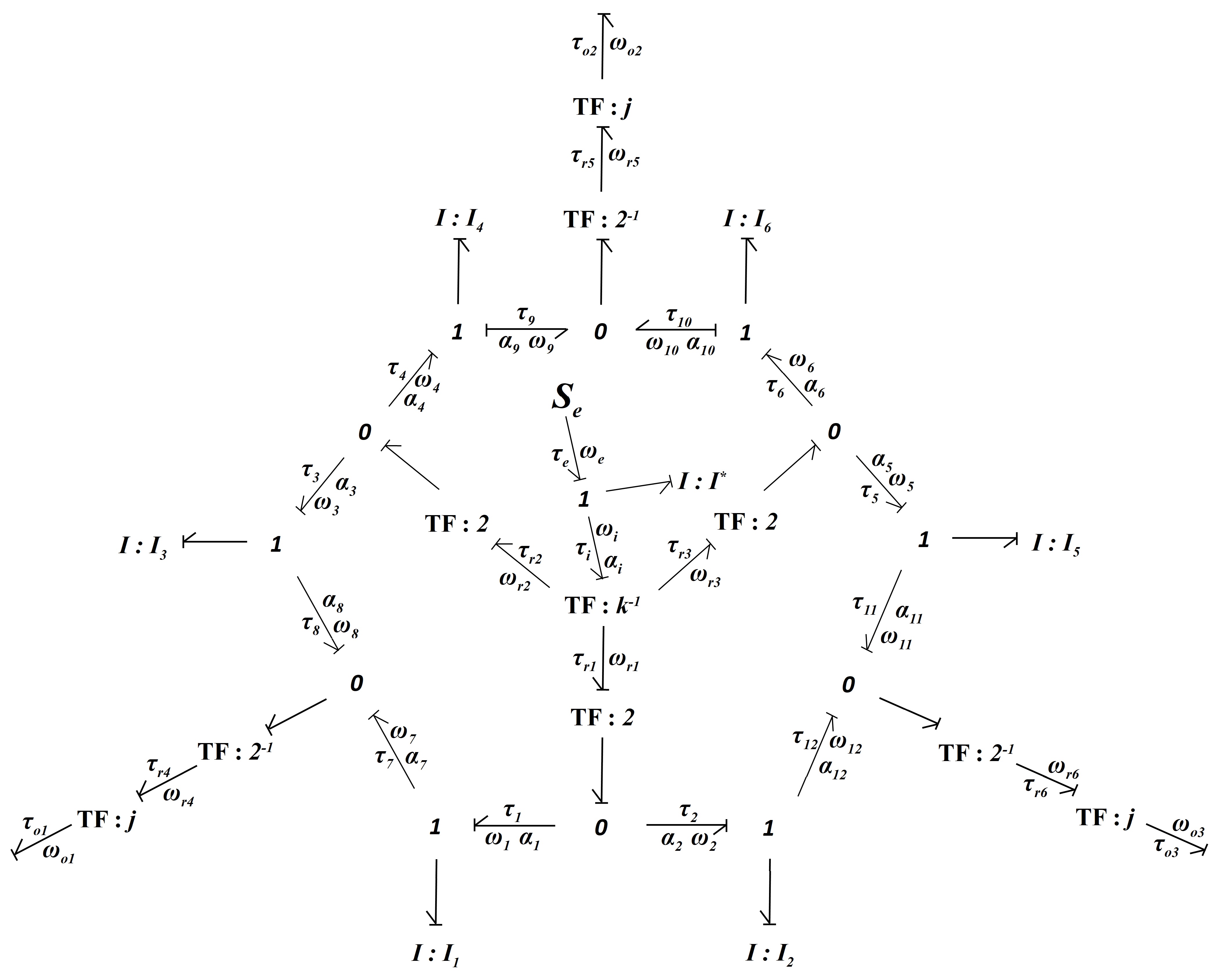}}
\caption{BONG GRAPH MODEL OF THE DIFFERENTIAL.}
\label{figure8}\vspace{-0.15in}
\end{figure}

The kinematics and the dynamics of the $3-OOD$ are derived by means of the bond graph modelling technique. The bond graph method is used as it enables a direct correlation for the mathematical model of the system with its physical framework \cite{deur2012bond}. The kinematic scheme $($kinematic diagram$)$ as shown in Fig.~\ref{figure7} illustrates the structure and the links between the components of the $3-OOD$. The bond graph model as shown in Fig.~\ref{figure8} is developed based on its corresponding kinematic scheme. All the parameters used in formulating the kinematic scheme and the bond graph model are listed in Table~\ref{table_ASME}.
\begin{table}[t]\vspace{0.075in}
\caption{PARAMETERS OF COMPONENTS OF $3-OOD$}\vspace{-0.15in}
\begin{center}
\label{table_ASME}
\begin{tabular}{c c c c c}
& & \\ 
\hline
Components & $\omega$ & $\alpha$ & Torque & Inertia \\
\hline
Source $(S_e)$ & $\omega _e$ & $\alpha _e$ & $\tau _e$
 \\
Input $(I)$ & $\omega _i$ & $\alpha _i$ & $\tau _i$ & $I_i$
\\
Output $(O_{1})$ & $\omega _{O1}$ & & $\tau _{O1}$
\\
Output $(O_{2})$ & $\omega _{O2}$ & & $\tau _{O2}$
\\
Output $(O_{3})$ & $\omega _{O3}$ & & $\tau _{O3}$
\\
Ring Gear $(R1)$ & $\omega _{R1}$ & $\alpha _{R1}$ & $\tau _{R1}$
\\
Ring Gear $(R2)$ & $\omega _{R2}$ & $\alpha _{R2}$ & $\tau _{R2}$
\\
Ring Gear $(R3)$ & $\omega _{R3}$ & $\alpha _{R3}$ & $\tau _{R3}$
\\
Ring Gear $(R4)$ & $\omega _{R4}$ & $\alpha _{R4}$ & $\tau _{R4}$
\\
Ring Gear $(R5)$ & $\omega _{R5}$ & $\alpha _{R5}$ & $\tau _{R5}$
\\
Ring Gear $(R6)$ & $\omega _{R6}$ & $\alpha _{R6}$ & $\tau _{R6}$
\\
Side Gear $(S_1)$ & $\omega _1$ & $\alpha _1$ & $\tau _1$
\\
Side Gear $(S_2)$ & $\omega _2$ & $\alpha _2$ & $\tau _2$
\\
Side Gear $(S_3)$ & $\omega _3$ & $\alpha _3$ & $\tau _3$
\\
Side Gear $(S_4)$ & $\omega _4$ & $\alpha _4$ & $\tau _4$
\\
Side Gear $(S_5)$ & $\omega _5$ & $\alpha _5$ & $\tau _5$
\\
Side Gear $(S_6)$ & $\omega _6$ & $\alpha _6$ & $\tau _6$
\\
Side Gear $(S_7)$ & $\omega _7$ & $\alpha _7$ & $\tau _7$ & $I_1$
\\
Side Gear $(S_8)$ & $\omega _8$ & $\alpha _8$ & $\tau _8$ & $I_3$
\\
Side Gear $(S_9)$ & $\omega _9$ & $\alpha _9$ & $\tau _9$ & $I_4$
\\
Side Gear $(S_{10})$ & $\omega _{10}$ & $\alpha _{10}$ & $\tau _{10}$ & $I_6$
\\
Side Gear $(S_{11})$ & $\omega _{11}$ & $\alpha _{11}$ & $\tau _{11}$ & $I_5$
\\
Side Gear $(S_{12})$ & $\omega _{12}$ & $\alpha _{12}$ & $\tau _{12}$ & $I_2$
\\
\hline
\vspace{-0.55in}
\end{tabular}
\end{center}
\end{table}

It can be inferred from the bond graph model that the ring gears $(R_{1-3})$ of $(2-OD_{1-3})$ rotate with equal angular velocities that is $(\frac{1}{k})$ times the angular velocity of the input $(I)$ and with equal torque that is $(\frac{k}{3})$ times torque at the input, where $(k^{-1})$ is the gear ratio of the input to the ring gears. For the current design it is chosen that $(k^{-1} = \frac{1}{20})$ because the worm gear arrangement at the input requires a low speed ratios for smooth transmission.
\vspace{-0.1in}
\begin{equation}
\omega _{R1}=\omega _{R2}=\omega _{R3}=\frac{\omega _i}{k},
\label{1velocities_R-i}
\end{equation}
\vspace{-0.4in}
\begin{equation}
\tau _{R1}=\tau _{R2}=\tau _{R3}=\frac{k(\tau _i)}{3},
\label{2Torque_R-i}
\end{equation}
where $\omega _{R1}$, $\omega _{R2}$ and $\omega _{R3}$ and $\tau _{R1}$, $\tau _{R2}$ and $\tau _{R3}$ are the respective angular velocities and torques of the ring gears ($R_1$, $R_2$ and $R_3$), $\omega _i$ and $\tau _i$ are the angular velocity and torque of the input $(I)$. The side gears $(S_1$ and $S_2)$ of $2-OD_1$ can operate at different speeds $\omega _1$ and $\omega _2$, while maintaining equal torques $(\tau _1=\tau _2)$ \cite{deur2010modeling}. Similarly, the torques of the corresponding side gears $(S_3$ and $S_4)$ and $(S_5$ and $S_6)$ of $2-OD_2$ and $2-OD_3$ are equal $(\tau _3=\tau _4)$,$(\tau _5=\tau _6)$. The angular velocity $\omega R_1$ of the ring gear $(R_1)$ is always the average of the angular velocities $\omega _1$ and $\omega _2$ of the side gears $(S_1$ and $S_2)$ and the sum of the torques of the side gears $\tau _1$ and $\tau _2$ is equal to the torque $\tau _{R1}$ of the ring gear \cite{deur2010modeling}.
\vspace{-0.15in}
\begin{equation}
    \tau _1=\tau _2=\tau _3 = \tau _4=\tau _5=\tau _6=\frac{\tau _{R1}}{2},
    \label{3Torque_S-R}
\end{equation}
\begin{equation}
   \omega _{R1}=\frac{(\omega _1+\omega _2)}{2}.
\label{4velocities_S1,2-R1}\vspace{-0.25in}
\end{equation}
Similarly,
\vspace{-0.1in}
\small
\begin{equation}
\begin{split}
& \omega _{R2}=\frac{(\omega _3+\omega _4)}{2},\quad \omega _{R3}=\frac{(\omega _5+\omega _6)}{2},\ \omega _{R4}=\frac{(\omega _7+\omega _8)}{2},\\ & \omega _{R5}=\frac{(\omega _9+\omega _{10})}{2},\quad \omega _{R6}=\frac{(\omega _{11}+\omega _{12})}{2},
\end{split}
\label{5velocities_S3-12-R2-6}\vspace{-0.15in}
\end{equation}
\normalsize
where $\omega _{R2}$, $\omega _{R3}$, $\omega _{R4}$, $\omega _{R5}$ and $\omega _{R6}$ are the respective angular velocities of the ring gears $(R_2$, $R_3$, $R_4$, $R_5$ and $R_6)$ and $\omega _3$, $\omega _4$, $\omega _5$, $\omega _6$, $\omega _7$, $\omega _8$, $\omega _9$, $\omega _{10}$, $\omega _{11}$ and $\omega _{12}$ are the angular velocities of the respective side gears $(S_3$, $S_4$, $S_5$, $S_6$, $S_7$, $S_8$, $S_9$, $S_{10}$, $S_{11}$ and $S_{12})$. Substituting Eqn.~(\ref{4velocities_S1,2-R1}) and Eqn.~(\ref{5velocities_S3-12-R2-6}) in Eqn.~(\ref{1velocities_R-i}) and Eqn.~(\ref{3Torque_S-R}) in Eqn.~(\ref{2Torque_R-i}),
\vspace{-0.15in}
\begin{equation}
\omega _i=\frac{k(\omega _1+\omega _2)}{2}=\frac{k(\omega _3+\omega _4)}{2}=\frac{k(\omega _5+\omega _6)}{2},
\label{6velocities_S1-6-I}\vspace{-0.4in}
\end{equation}
\begin{equation}
\tau _i=\frac{6(\tau _1)}{k}.
\label{7Torque_S-i}
\end{equation}
It can be inferred from the bond graph model in Fig.~\ref{figure8} that the inertia $I_i$ of the input $(I)$ has a derived causality.
\vspace{-0.15in}
\begin{equation}
I_i\alpha _i = \tau _e-\tau _i,
\label{8Inertia_Ii}\vspace{-0.15in}
\end{equation}
where $\tau _e$ and $\tau _i$ are the respective torque at the source $S_e$ and the input $(I)$ and $\alpha _i$ is the angular acceleration of the input Fig.~\ref{figure8} similarly,
\vspace{-0.15in}
\begin{equation}
\begin{split}
&I_1\alpha _7 = \tau _1-\tau _7,\quad I_2\alpha _{12} = \tau _2-\tau _{12},\quad I_3\alpha _8 = \tau _3-\tau _8, \\ & I_4\alpha _9 = \tau _4-\tau _9,\quad I_5\alpha _{11} = \tau _5-\tau _{11},\quad I_6\alpha _{10} = \tau _6-\tau _{10},
\end{split}
\label{9Inertia_I1-6}\vspace{-0.25in}
\end{equation}
where $I_1$,$I_2$,$I_3$,$I_4$,$I_5$ and $I_6$ are the respective inertia exhibited by the side gear $(S_7$, $S_{12}$, $S_8$, $S_9$, $S_{11}$ and $S_{10})$ and $\alpha _7$, $\alpha _{12}$, $\alpha _8$, $\alpha _9$, $\alpha _{11}$ and $\alpha _{10}$ are their respective angular accelerations. The side gears $(S_1$ and $S_7)$ are connected such that there exists no relative motion between them.
\vspace{-0.2in}
\begin{equation}
\omega _1=\omega _7,
\label{10velocity_S1-S7}
\end{equation}
\vspace{-0.5in}
\begin{equation}
\alpha _1=\alpha _7.
\label{11accelration_S1-S7}\vspace{-0.15in}
\end{equation}
Similarly,
\small
\begin{equation}
\omega _2=\omega _{12},\quad \omega _3=\omega _8,\quad \omega _4=\omega _9,\quad \omega _5=\omega _{11},\quad \omega _6=\omega _{10},
\label{11velocity_S2-S12}
\end{equation}
\normalsize
\small
\begin{equation}
\alpha _2=\alpha _{12},\quad \alpha _3=\alpha _8,\quad \alpha _4=\alpha _9,\quad \alpha _5=\alpha _{11},\quad \alpha _6=\alpha _{10}.
\label{13accelration_S2-S12}
\end{equation}
\normalsize
Substituting Eqn.~(\ref{10velocity_S1-S7}) and Eqn.~(\ref{11velocity_S2-S12}) in the Eqn.~(\ref{6velocities_S1-6-I}),
\vspace{-0.15in}
\begin{equation}
\begin{split}
&\omega _7=\omega _1=\frac{2(\omega _i)}{k}-\omega _2,\qquad \omega _8=\omega _3=\frac{2(\omega _i)}{k}-\omega _4, \\ &\omega _{11}=\omega _5=\frac{2(\omega _i)}{k}-\omega _6.
\end{split}
\label{12velocity_S1-12-I}\vspace{-0.15in}
\end{equation}

\subsection*{The Input To Outputs Kinematics}
\vspace{0.0in}

Similar to Eqn.~(\ref{1velocities_R-i}) the angular velocities $\omega _{O1}$, $\omega _{O2}$ and $\omega _{O3}$ of the respective outputs $(O_1$, $O_2$ and $O_3)$ is $j$ times the respective velocities $\omega _{R4}$, $\omega _{R5}$ and $\omega _{R5}$ of the corresponding ring gears $(R_4$, $R_5$ and $R_6)$, where $j$ is the gear ratio of the ring gears $(R_{4-6})$ to the outputs $(O_{1-3})$. For the current design it is chosen that $(j = \frac{2}{1})$.
\vspace{-0.15in}
\begin{equation}
\omega _{O1}=j(\omega _{R4}),\quad \omega _{O2}=j(\omega _{R5}),\quad \omega _{O3}=j(\omega _{R6}).
\label{13velocity_O-R}\vspace{-0.15in}
\end{equation}
Substituting Eqn.~(\ref{5velocities_S3-12-R2-6}) in Eqn.~(\ref{13velocity_O-R}),
\vspace{-0.1in}
\begin{equation}
\begin{split}
    &\omega _{O1}=\frac{j(\omega _7+\omega _8)}{2},\qquad \omega _{O2}=\frac{j(\omega _9+\omega _{10})}{2}, \\ & \omega _{O3}=\frac{j(\omega _{11}+\omega _{12})}{2}.
\end{split}
\label{14velocity_O-S7-12}\vspace{-0.3in}
\end{equation}
Further substituting Eqn.~(\ref{11velocity_S2-S12}) and Eqn.~(\ref{12velocity_S1-12-I}) in Eqn.~(\ref{14velocity_O-S7-12}),
\vspace{-0.15in}
\begin{equation}
\omega _{O1}=\frac{2j(\omega_i)}{k}-\frac{j(\omega _2+\omega _4)}{2},
\label{15velocity_O1-I}\vspace{-0.3in}
\end{equation}
\begin{equation}
\omega _{O2}=\frac{2j(\omega_i)}{k}-\frac{j(\omega _3+\omega _5)}{2},
\label{16velocity_O2-I}\vspace{-0.3in}
\end{equation}
\begin{equation}
\omega _{O3}=\frac{2j(\omega _i)}{k}-\frac{j(\omega _1+\omega _6)}{2}.
\label{17velocity_O3-I}\vspace{-0.15in}
\end{equation}
\textbf{\small Equivalent kinematics of the outputs:\normalsize} Equations (\ref{15velocity_O1-I}), (\ref{16velocity_O2-I}) and (\ref{17velocity_O3-I}) show that all the three outputs have equivalent kinematics with the input. The outputs $(O_{1-3})$ relate to the input $(I)$ with additional dependence on the side gears $(S_{1-6})$. This is because the $3-OOD$ has three degrees of freedom (3DOF) and the behaviour of an output is impacted by the input $(I)$ and the other two outputs. In Eqn.~(\ref{15velocity_O1-I}) the influence of the outputs $(O_{2}$ and $O_{3})$ on the output $(O_{1})$ is seen in the form of $\omega _2$ and $\omega _4$.
\subsection*{The Input To Outputs Dynamics}
\vspace{0.0in}

Similar to Eqn.~(\ref{2Torque_R-i}), the torque $\tau_{O1}$, $\tau_{O2}$ and $\tau_{O3}$ of the respective outputs $(O_1$, $O_2$ and $O_3)$ is ($j^{-1}$) times the respective torque $\tau_{R4}$, $\tau_{R5}$ and $\tau_{R6}$ of the corresponding ring gears $(R_4$, $R_5$ and $R_6)$.
\vspace{-0.2in}
\begin{equation}
\tau_{O1}=\frac{\tau_{R4}}{j},\quad \tau_{O2}=\frac{\tau_{R5}}{j},\quad \tau_{O3}=\frac{\tau_{R6}}{j}.
\label{18torque_O-R4-6}\vspace{-0.15in}
\end{equation}
It can be inferred from the bond graph model in Fig.~\ref{figure8} that the torque $\tau_{R4}$ is the sum of the torques $\tau_7$ and $\tau_8$ of the side gears $S_7$ and $S_8$ of $2-ID_1$ and similarly for $\tau_{R5}$ and $\tau_{R6}$
\vspace{-0.15in}
\begin{equation}
\tau_{R4}=\tau_7+\tau_8,\quad \tau_{R5}=\tau_9+\tau_{10},\quad \tau_{R6}=\tau_{11}+\tau_{12}.
\label{19torque_R4-6-S7-12}\vspace{-0.15in}
\end{equation}
Substituting Eqn.~(\ref{19torque_R4-6-S7-12}) in Eqn.~(\ref{18torque_O-R4-6}),
\small
\begin{equation}
\tau_{O1}=\frac{(\tau_7+\tau_8)}{j},\quad \tau_{O2}=\frac{(\tau_9+\tau_{10})}{j},\quad \tau_{O3}=\frac{(\tau_{11}+\tau_{12})}{j}.
\label{20torque_O-S7-12}
\end{equation}
\normalsize
Further substituting Eqn.~(\ref{3Torque_S-R}), Eqn.~(\ref{7Torque_S-i}) and Eqn.~(\ref{9Inertia_I1-6}) in Eqn.~(\ref{20torque_O-S7-12}),
\vspace{-0.15in}
\begin{equation}
\tau_{O1}=\frac{k(\tau_i)}{3j}-\frac{(I_1\alpha _7+I_3\alpha _8)}{j},
\label{21torque_O1-I}\vspace{-0.4in}
\end{equation}
\begin{equation}
\tau_{O2}=\frac{k(\tau_i)}{3j}-\frac{(I_4\alpha _9+I_6\alpha _{10})}{j},
\label{22torque_O2-I}\vspace{-0.4in}
\end{equation}
\begin{equation}
\tau_{O3}=\frac{k(\tau_i)}{3j}-\frac{(I_2\alpha _{12}+I_5\alpha _{11})}{j}.
\label{23torque_O3-I}\vspace{-0.15in}
\end{equation}
\textbf{\small Equivalent dynamics of the outputs:\normalsize} Equations (\ref{21torque_O1-I}), (\ref{22torque_O2-I}) and (\ref{23torque_O3-I}) show that all the three outputs of the differential have equivalent dynamics with the input.
\subsection*{When Three Outputs Are Under Equal Loads}
\vspace{0.0in}
Since the side gears $(S_{7-12})$ are all identical, they exhibit equal inertia. Furthermore, when all three outputs $(O_{1-3})$ are under equal loads (resistive torque `$\tau_R$'), the side gears $(S_{7-12})$ operate with equal angular velocity and angular acceleration. 
\vspace{-0.15in}
\begin{equation}
\begin{split}
    &\omega _1=\omega _2=\omega _3=\omega _4=\omega _5=\omega _6= \\ & \omega _7=\omega _8=\omega _9=\omega _{10}=\omega _{11}=\omega _{12}=\frac{\omega _i}{k},
\end{split}
\label{24velocity_S1-12-I}
\end{equation}\vspace{-0.4in}
\begin{equation}
\begin{split}
    &\alpha _1=\alpha _2=\alpha _3=\alpha _4=\alpha _5=\alpha _6= \\ & \alpha _7=\alpha _8=\alpha _9=\alpha _{10}=\alpha _{11}=\alpha _{12},
\end{split}
\label{27accelration_S1-12}
\end{equation}\vspace{-0.4in}
\begin{equation}
I_1=I_2=I_3=I_4=I_5=I_6.
\label{25Inertia_S7-12}\vspace{-0.15in}
\end{equation}
Substituting Eqn.~(\ref{6velocities_S1-6-I}), Eqn.~(\ref{24velocity_S1-12-I}) in Eqn.~(\ref{15velocity_O1-I}), Eqn.~(\ref{16velocity_O2-I}) and Eqn.~(\ref{17velocity_O3-I}),
\vspace{-0.2in}
\begin{equation}
{\bf\omega _{O1}}={\bf\omega _{O2}}={\bf\omega _{O3}}=\frac{j(\omega _i)}{k}.
\label{newVelocity_O1=O2=O3}\vspace{-0.15in}
\end{equation}
Substituting Eqn.~(\ref{27accelration_S1-12}) and Eqn.~(\ref{25Inertia_S7-12}) in Eqn.~(\ref{21torque_O1-I}), Eqn.~(\ref{22torque_O2-I}) and Eqn.~(\ref{23torque_O3-I}) and subtracting the resistive torque $(\tau_R)$ from the torque of each output
\vspace{-0.2in}
\begin{equation}
{\bf\tau _{O1}}={\bf\tau _{O2}}={\bf\tau _{O3}}=\frac{k(\tau_i)}{3j} - \frac{2(I_1\dot{\omega}_1)}{j} - \tau_R,
\label{newTorque_O1=O2=O3}\vspace{-0.15in}
\end{equation}
\textbf{\small Equivalent angular velocities and torque :\normalsize} Equations (\ref{newVelocity_O1=O2=O3}) and (\ref{newTorque_O1=O2=O3}) illustrate the novel ability of the differential to translate equal motion $(\omega _{O1}=\omega _{O2}=\omega _{O3})$ and equal torque $(\tau _{O1}=\tau _{O2}=\tau _{O3})$ to all its three outputs that are unconstrained $(\tau_R = 0)$ or under equal loads.

\subsection*{Kinematics Between The Outputs}
\vspace{0.0in}

The kinematics an output of the $3-OOD$ shares with the other two outputs is determined by substituting Eqn.~(\ref{11velocity_S2-S12}) and Eqn.~(\ref{14velocity_O-S7-12}) into the previously established input to outputs kinematics in Eqn.~(\ref{15velocity_O1-I}).
\vspace{-0.15in}
\begin{equation}
\omega _{O1}=\frac{2j(\omega _i)}{k}+\frac{j(\omega _{10}+\omega _{11})}{2} -(\omega _{O2}+\omega _{O3}),
\label{29Velocity_O1--O2,O3}\vspace{-0.15in}
\end{equation}
Further substituting Eqn.~(\ref{11velocity_S2-S12}) and Eqn.~(\ref{6velocities_S1-6-I}) in Eqn.~(\ref{29Velocity_O1--O2,O3}),
\vspace{-0.15in}
\begin{equation}
\omega _{O1}+\omega _{O2}+\omega _{O3}=\frac{3j(\omega _i)}{k}.
\label{30Velocity_O1-O2,O3}\vspace{-0.15in}
\end{equation}

\subsection*{Dynamics Between The Outputs}

The dynamics an output of the $3-OOD$ shares with the other two outputs is determined by substituting Eqn.~(\ref{9Inertia_I1-6}), Eqn.~(\ref{11accelration_S1-S7}) and Eqn.~(\ref{13accelration_S2-S12}) into the previously determined input to outputs dynamics in Eqn.~(\ref{21torque_O1-I}).
\vspace{-0.15in}
\begin{equation}
\tau _{O1}=\frac{2k(\tau_i)}{3j}- \frac{(I_1\alpha _7+I_3\alpha _8)}{j}-\frac{(\tau _1+\tau _3)}{j}.
\label{31Torque_O1--O2,O3}\vspace{-0.15in}
\end{equation}
Substituting Eqn.~(\ref{3Torque_S-R}) and Eqn.~(\ref{9Inertia_I1-6}) in Eqn.~(\ref{31Torque_O1--O2,O3}),
\small
\begin{equation}
\begin{split}
    \tau _{O1}= \frac{2k(\tau_i)}{3j} - \frac{(I_1\alpha _7+I_2\alpha _{12}+I_3\alpha _8+I_4\alpha _9)}{j} - \frac{(\tau _9+\tau _{12})}{j}.
\end{split}
\label{32Torque_O1--O2,O3}\vspace{-0.15in}
\end{equation}
\normalsize
Further substituting Eqn.~(\ref{9Inertia_I1-6}), Eqn.~(\ref{13accelration_S2-S12}) and Eqn.~(\ref{20torque_O-S7-12}) in Eqn.~(\ref{32Torque_O1--O2,O3}),
\small
\vspace{-0.1in}
\begin{equation}
    \begin{split}
        \tau _{O1}+\tau _{O2}+&\tau _{O3}= \frac{k(\tau_i)}{j} \\ & -\frac{(I_1\alpha _7+I_2\alpha _{12}+I_3\alpha _8 + I_4\alpha _9+I_5\alpha _{11}+I_6\alpha _{10})}{j}.
    \end{split}
    \label{33Torque_O1-O2,O3}
\end{equation}
\normalsize
When motion and torque are provided to one of the outputs, the input receives no motion $(\omega_i=0$, $\tau_i=0)$ because the worm gear arrangement at the input only allows one-way motion from the input to the ring gears. The motion provided to the outputs is shared between the other two outputs which will rotate in the opposite direction. Substituting $(\omega_i=0)$ in Eqn.~(\ref{30Velocity_O1-O2,O3}), $(\tau_i=0)$ in Eqn.~(\ref{33Torque_O1-O2,O3}),
\vspace{-0.15in}
\begin{equation}
\omega_{O1}=-(\omega _{O2} + \omega _{O3}),
\label{34Velocity_O1=-O2+O3}
\end{equation}
\vspace{-0.2in}
\small
\begin{equation}
    \begin{split}
        \tau _{O1}=& - \tau _{O2} - \tau _{O3}  \\ & -\frac{(I_1\alpha _7+I_2\alpha _{12}+I_3\alpha _8 + I_4\alpha _9+I_5\alpha _{11}+I_6\alpha _{10})}{j}.
    \end{split}
    \label{35Torque_O1-O2,O3}
\end{equation}
\normalsize
The negative sign indicates that $(O_2)$ and $(O_3)$ move in the opposite direction of $(O_1)$.

\textbf{\small Novelty of \textit{3-OOD}:\normalsize} Equations (\ref{newVelocity_O1=O2=O3}) and (\ref{newTorque_O1=O2=O3}) illustrate the novel ability of the differential to translate equal motion $(\omega _{O1}=\omega _{O2}=\omega _{O3})$ and equal torque $(\tau _{O1}=\tau _{O2}=\tau _{O3})$ to all its three outputs that are unconstrained $(\tau_R = 0)$ or under equal loads. Equations (\ref{30Velocity_O1-O2,O3}) and (\ref{33Torque_O1-O2,O3}) show the equivalent kinematic and dynamics relations that the outputs of the differential share with each other. Furthermore, Eqn.~(\ref{34Velocity_O1=-O2+O3}) and Eqn.~(\ref{35Torque_O1-O2,O3}) illustrate that when the input is ceased from motion and motion is provided to one of the outputs, the two other outputs receive equal motion when under equal loads. In such cases the $3-OOD$ behaves as a $2-OD$ where the output which receives motion acts as the input and the other two outputs will act as the outputs of a $2-OD$.

\section*{CURRENT AND POTENTIAL APPLICATIONS}
\vspace{0.0in}
\begin{figure} 
\centerline{\includegraphics[width=3.25in]{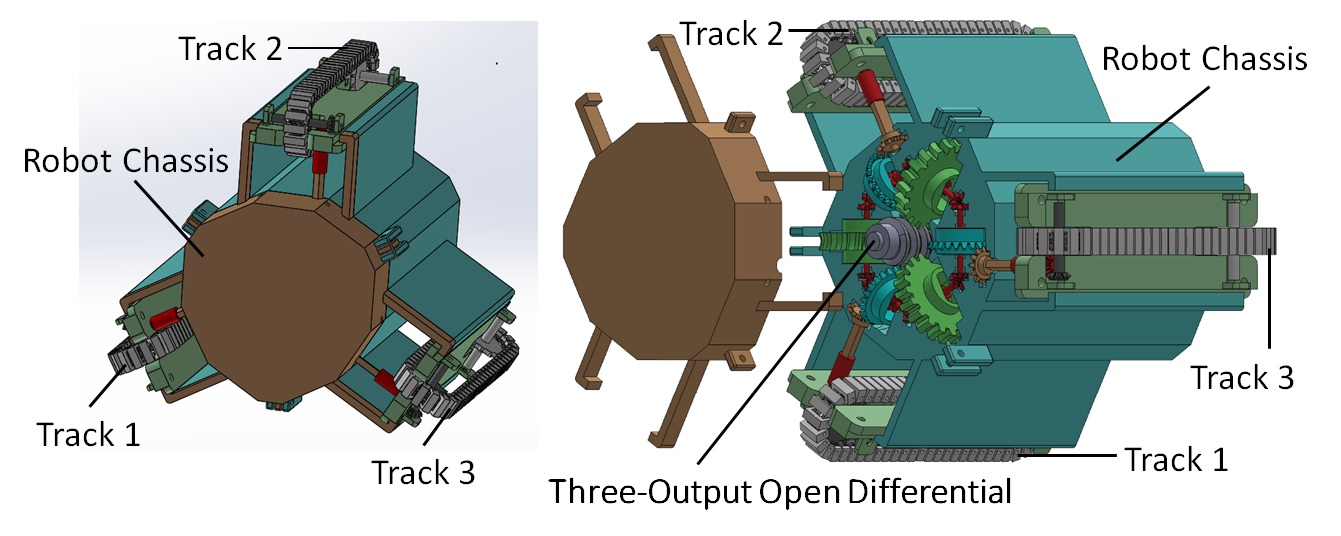}}
\caption{$3-OOD$ INSIDE A PIPE CLIMBER.}
\label{figure14}
\end{figure}
The $3-OOD$ is designed to be used inside a pipe climbing robot as shown in Fig.~\ref{figure14} \cite{vadapalli2021modular,kumar2021design}. It can equip the three tracks of the robot with differential speed so that, during a turn it can rotate the track travelling the longer distance faster than the track that travels the shorter distance. But when moving in a straight path the differential can rotate the thee tracks with equal speeds. Using the differential in a pipe climber will potentially eliminate slip and drag of its tracks and reduce the stresses the robot experiences during turns and thus providing a smoother motion of the robot \cite{vadapalli2021modular}. Other current applications can be found in robotic arms such as the SARAH (Self-Adapting Robotic Auxilary Hand) $($Self-Adapting Robotic Auxilary Hand$)$ and sensor less in-hand manipulation by underactuated robot hand \cite{ospina2020sensorless,laliberte2001underactuation,wang2016underactuated}. The differential can also find applications in single-input multiple-spindle wrench $($nut-runner$)$ as suggested by S. Kota and S. Bidare (1997) \cite{kota1997systematic}. Other potential applications for the differential include but not limited manipulators, legged robots, and wheeled robots \cite{yin2020research}. In the near future, a prototype of the differential will be made to further research the functioning of the robot under different load cases and also study the transmission efficiency and energy flow of the differential. In addition, we will be implementing the differential in a pipe climbing robot.

\section*{CONCLUSION}
\vspace{0.0in}

The novel design of the $3-OOD$ is presented. $3-OOD$ is designed such that its functioning ability is analogous to the traditional $2-OD$. The kinematics and dynamics of the differential are established using the bond graph model. From the analytical and theoretical study conducted it is observed that the $3-OOD$ is the first differential with three-outputs to realise all the functioning abilities of a traditional $2-OD$. That is, $(i)$ its outputs operate with equal speed when the outputs are unconstrained or under equal loads, $(ii)$ its input to output kinematics and dynamics is equivalent for all three outputs and $(iii)$ its kinematics and dynamics between any two of its outputs is equivalent. It is also observed that when one of the outputs is ceased from motion and the other two outputs are left unconstrained, they operate with equal angular velocities and equal torques. The need for such a mechanism is discussed with relevant current and potential applications in fields such as robotics and automobile engineering.

\bibliographystyle{asmems4}
%
% ...

\bibliography{asme2e}

\end{document}